\documentclass{article}

\usepackage{microtype}
\usepackage{graphicx}
\usepackage{subcaption}
\usepackage{booktabs} 

\usepackage{hyperref}

\usepackage[preprint]{icml2026}

\usepackage{amsmath}
\usepackage{amssymb}
\usepackage{mathtools}
\usepackage{amsthm}

\usepackage[table]{xcolor}

\newcommand{\eg}{\emph{e.g.}}

\usepackage[capitalize,noabbrev]{cleveref}

\usepackage[raster,skins, most]{tcolorbox} %

\definecolor{purple1}{RGB}{126, 107, 196}

\theoremstyle{plain}

\theoremstyle{definition}

\theoremstyle{remark}

\usepackage[textsize=tiny]{todonotes}

\icmltitlerunning{
VSearcher: Long-Horizon Multimodal Search Agent via Reinforcement Learning
}

\begin{document}

\twocolumn[
  \icmltitle{
  VSearcher: Long-Horizon Multimodal Search Agent \\ via Reinforcement Learning
  }



  \icmlsetsymbol{equal}{*}




  \begin{icmlauthorlist}
    \icmlauthor{Ruiyang Zhang}{um,idea}
    \icmlauthor{Qianguo Sun}{idea}
    \icmlauthor{Chao Song}{idea}
    \icmlauthor{Yiyan Qi}{idea}
    \icmlauthor{Zhedong Zheng}{um}

    \vspace{20pt}
    \small{\url{https://github.com/Ruiyang-061X/VSearcher}}
  \end{icmlauthorlist}

  \icmlaffiliation{um}{FST and ICI, University of Macau, China}
  \icmlaffiliation{idea}{IDEA}

  \icmlcorrespondingauthor{Zhedong Zheng}{zhedongzheng@um.edu.mo}

  \icmlkeywords{Machine Learning, ICML}

  \vskip 0.3in
]



\printAffiliationsAndNotice{}  

\begin{abstract}
Large models are increasingly becoming autonomous agents that interact with real-world environments and use external tools to augment their static capabilities.
However, most recent progress has focused on text-only large language models, which are limited to a single modality and therefore have narrower application scenarios. On the other hand, multimodal large models, while offering stronger perceptual capabilities, remain limited to static knowledge and lack the ability to access and leverage up-to-date web information.
In this paper, we propose VSearcher, turning static multimodal model into multimodal search agent capable of long-horizon, multi-turn tool use in real-world web environments, including text search, image search, and web browsing, via reinforcement learning. Specifically, we introduce Iterative Injection Data Synthesis pipeline to generate large-scale, complex multimodal QA questions, which are further filtered with comprehensive metrics to ensure high quality and sufficient difficulty. We then adopt an SFT-then-RL training pipeline to turn base multimodal models to agent capable of multi-turn tool calling in real-world web environments. Besides, we propose a multimodal search benchmark MM-SearchExam dedicated to evaluating search capabilities of multimodal search agents, which proves highly challenging for recent proprietary models.
Extensive evaluations across multiple multimodal search benchmarks reveal effectiveness of our method. VSearcher achieves superior performance compared to recent multimodal search agents and even surpasses several proprietary models on multimodal web search tasks.
\end{abstract}

\section{Introduction}

Since the introduction of strong reasoning models such as OpenAI o1~\cite{jaech2024openai} and DeepSeek-R1~\cite{guo2025deepseek}, foundation models are increasingly equipped with inherent step-by-step reasoning capabilities. Moreover, with paradigms such as ReAct~\cite{yao2022react}, these reasoning models are developed into agents that iteratively think and act to break down complex problems. Notably, large models are increasingly becoming autonomous agents that can interact with real-world environments and harness external tools to augment their static, fixed knowledge~\cite{li2025websailor,tao2025webleaper}.
However, recent progress in building autonomous agents has primarily focused on large language models~\cite{ferrag2025llm,lin2025creativity,luo2025large}. This single-modality constraint greatly limits their application scenarios and prevents them from meeting real-world user needs. In contrast, multimodal large models~\cite{bai2025qwen2,li2024llava,wu2023multimodal,zhang2026sketchthinker}, with their stronger perceptual abilities, often remain restricted to static knowledge and cannot effectively harness external tools to access up-to-date information from real-world web environments.

In this paper, we propose VSearcher, transforming static multimodal models into autonomous agents capable of long-horizon, multi-turn tool use in real-world web environments, including image search, text search, and web browsing, via reinforcement learning. Notably, our model is trained in real-world web environments with challenging multimodal tasks, resulting in agent that can navigate extensive web sources and solve real-world multimodal user requests.
Specifically, we propose a comprehensive post-training framework for building multimodal search agents, spanning data synthesis, rejection fine-tuning, and reinforcement learning. Moreover, we introduce a challenging multimodal browsing benchmark to enable more thorough evaluation of multimodal search agents.
For data synthesis, we propose Iterative Injection-based Data Synthesis to fully automatically generate large-scale, challenging multimodal browsing tasks. First, we obtain rare seed entities from the Wikidata database using tailored SPARQL rules to ensure rarity. Next, we collect the content of the corresponding Wikipedia pages and construct an initial, simple text QA pair from the content. We then select an entity from the question, retrieve Wikipedia information about this entity, and transform the original question by hiding the selected entity and replacing it with parsed information; this process is iteratively applied for several rounds. Finally, we select an entity that is critical for answering the question, retrieve an image of this entity, and hide the entity in the question with a phrase such as `shown in the image'; the image then serves as the visual input for the question. We also propose several filtering metrics to ensure the difficulty and quality of the synthesized questions.
For rejection fine-tuning, we aim to instill strong browsing ability from powerful teacher model into base multimodal model to provide an initial long-horizon tool-use capability. Specifically, we build a ReAct~\cite{yao2022react} inference framework and utilize Gemini-3-Pro-Thinking as the teacher model. We generate trajectories for our synthesized multimodal browsing questions with Gemini-3-Pro-Thinking and reject those trajectories that end with incorrect answers. Finally, we fine-tune base multimodal models on the remaining high-quality trajectories.
For reinforcement learning, we perform GRPO~\cite{shao2024deepseekmath} on the fine-tuned model to further generalize long-horizon search ability. Notably, the entire training process is conducted in real web environments to enhance real-world browsing performance. We utilize final-answer correctness as the trajectory reward. Strict format checking is enabled during rollout to ensure correct formatting throughout the whole rollout process. We utilize our synthesized questions as the training data for the reinforcement learning stage.
For the benchmark, we curate highly challenging questions from our data synthesis pipeline. We apply more extensive iterative generation process to ensure the difficulty of these questions. Several recent proprietary models achieve low accuracy on our benchmark.

Thorough evaluation on several multimodal search benchmarks validates the effectiveness of our trained model. VSearcher surpasses recent multimodal browsing agents and even exceeds the performance of several proprietary models.
In summary, our contributions are as follows:

\begin{itemize}
    \item \textbf{Date synthesis pipeline for multimodal browsing task.} We propose Iterative Injection-based Data Synthesis, together with comprehensive filtering metrics, enabling fully automatic, large-scale generation of challenging multimodal browsing tasks.
    \item \textbf{A comprehensive post-training pipeline for multimodal search agent.} We propose a thorough post-training pipeline for transforming base multimodal model into long-horizon multimodal search agent. Rejection sampling fine-tuning instills initial multi-turn tool-use capability, and reinforcement learning generalizes long-horizon search ability in real-world web environments.
    \item \textbf{Challenging benchmark for evaluation of multimodal search agent.} We curate a challenging benchmark MM-SearchExam for evaluating multimodal browsing ability, on which strong proprietary models achieve low performance.
    \item \textbf{Strong performance on various multimodal browsing benchmark} VSearcher achieves stronger results on several multimodal search benchmarks than recent multimodal browsing agents, and surpasses performance of several proprietary models.
\end{itemize}

\begin{figure*}
    \centering
    \includegraphics[width=0.9\linewidth]{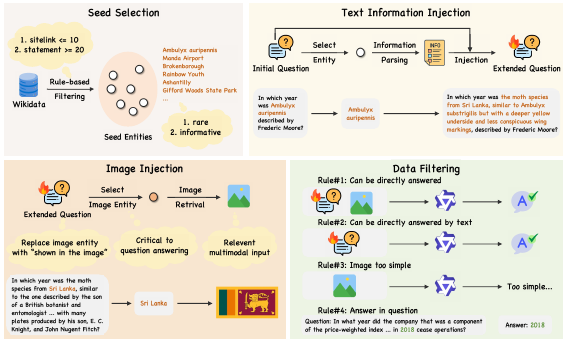}
    \vspace{-5pt}
    \caption{
    Our proposed Iterative Injection-based Data Synthesis.
    }
    \label{fig:data_synthesis}
    \vspace{-10pt}
\end{figure*}

\section{Method}

In this section, we provide an overview of our post-training framework for VSearcher, including Iterative Injection-based Data Synthesis, Rejection Sampling Fine-tuning, and Reinforcement Learning. We also describe the curation details of our proposed multimodal search benchmark, MM-SearchExam.

\subsection{Iterative Injection-based Data Synthesis.}

Challenging browsing tasks form the foundation for training and building long-horizon multimodal search agents. We propose Iterative Injection-based Data Synthesis, together with comprehensive data-filtering metrics, to synthesize challenging multimodal browsing tasks fully automatically. In summary, we begin with a text-only seed entity and iteratively extend it into complex multimodal QA questions (see Fig.~\ref{fig:data_synthesis}).

\noindent \textbf{Seed Selection.} Seed selection is critical to the difficulty of the final synthesis questions. Seeds for browsing tasks should be as rare as possible to encourage the browsing agent to rely on external tools rather than its static knowledge. We utilize Wikidata as the data source, which serves as the structured knowledge base behind Wikipedia. In Wikidata, a `sitelink' refers to the links of an entity across different languages; fewer sitelinks typically indicate a rarer entity. A `statement' denotes factual triples associated with an entity; more statements usually imply that the corresponding Wikipedia page contains richer information. During seed selection, we explicitly select entities with sitelinks below a threshold and statements above a threshold, yielding seeds that are both rare and sufficiently documented for subsequent iterative extension. We use SPARCLE to perform rule-based filtering over Wikidata to obtain seed entities satisfying our requirements.

\noindent \textbf{Initial QA Generation.} For each selected seed, we parse its Wikipedia content from an offline Wikipedia index. Due to the design of the seed selection stage, the corresponding content tends to be both rarely-known and sufficiently detailed. We then prompt an off-the-shelf LLM to generate an initial text-only QA pair from the content. We instruct the LLM to generate a one-sentence question about the provided entity. The question should be specific enough to have a unique answer. The generated answer should be a single word or a short phrase, making correctness easy to verify.

\noindent \textbf{Text Information Injection.} After obtaining the initial QA pair, we perform multiple rounds of text information injection to gradually extend the initial question into a complex browsing problem. In each round, we select an entity from the question, hide it, and replace it with its parsed information. This process increases question complexity and the need for web browsing. Specifically, in each round, we first prompt an LLM to select an entity from the initial question and obtain the corresponding Wikipedia content of the selected entity. We then prompt an LLM to extract one specific piece of information from the Wikipedia content. Notably, we instruct the LLM to extract rarely known information rather than commonly known facts, as the latter are less effective at increasing difficulty. Moreover, we encourage the LLM to include as many entities as possible in the extracted information, which facilitates subsequent rounds of iterative injection by providing additional entities to extend. Finally, we prompt an LLM to transform the original question by hiding the selected entity, replacing it with the parsed information, and ensuring that the transformed question remains logically coherent.

\noindent \textbf{Image Injection.} After several rounds of text information injection, the initial QA pair becomes a complex browsing problem. We then perform image injection to turn the extended question into multimodal question. For image injection, we first prompt an LLM to select an entity in the question that is critical for correct answering of the question. We then retrieve the Wikipedia image of the selected entity. The retrieved image serves as the input image for the question. Next, we prompt an LLM to transform the question by replacing the entity mention with a phrase such as `shown in the image'. After multiple rounds of text information injection, the question tends to be complex and the selected image entity tends to be rare. As a result, the image of this entity also tends to be rare, which encourages the model to call image search tool to obtain additional information of the image. Moreover, because we choose an entity that is crucial for answering the question and replace it with its image, the image plays a larger role in solving the question, avoiding cases where the model can solve the browsing task solely from text.

\noindent \textbf{Difficulty Level Design.} To train a multimodal browsing agent that is proficient across tasks with varying difficulty and browsing requirements, we generate multimodal browsing questions at different difficulty levels and mix them during training. This strategy is beneficial for building an agent with adaptive and robust browsing skills. Specifically, we synthesize 3 levels of browsing tasks: easy, medium, and hard. We conduct 1 round, 3 rounds, and 5 rounds of text information injection for easy, medium, and hard browsing tasks respectively.

\noindent \textbf{Data Filtering Strategy.} We design comprehensive filtering metrics for our synthesized multimodal browsing tasks to ensure both high difficulty and high quality. (1) We filter out questions that can be answered directly by a weaker LVLM. Even with multiple rounds of iterative extension, the combination of injected information could create reasoning shortcuts that degrade the task to simple QA question. We remove such samples because they typically lead to no tool use during subsequent rejection sampling fine-tuning and reinforcement learning, and therefore do not improve long-horizon tool-use capability. (2) We filter out samples that can be answered directly from the textual content by a weaker LLM. Although we explicitly select an entity that is critical for answering the question during image injection, there are still cases where the answer can be inferred from text alone. In such samples, the multimodal information plays a minor role, so we remove them because they are less beneficial for improving multimodal perception ability in browsing agents. (3) We filter out samples with too simple image. During case studies, we observe image that contains little information and are overly simple, \eg, the flag of a commonly known country. Such samples are not beneficial for improving multimodal browsing skills such as calling image search, extracting information from image, and perform text search based on those information to obtain more context. (4) We filter out samples where the answer is already present in the question. These occur as corner cases: during information injection, the extracted information could accidentally include the ground-truth answer and be injected into the question. We remove these corner cases.

\begin{figure*}
    \centering
    \includegraphics[width=0.8\linewidth]{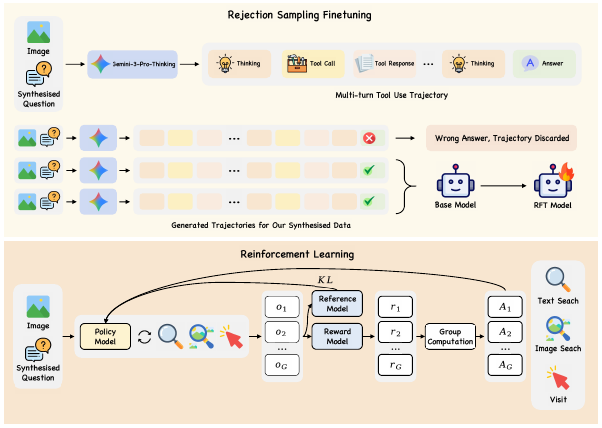}
    \vspace{-5pt}
    \caption{
    Rejection sampling finetuning and reinforcement learning process for VSearcher.
    }
    \label{fig:method}
    \vspace{-10pt}
\end{figure*}

\subsection{Rejection Sampling Finetuning}

In this stage, we instill strong multi-turn browsing skills from proprietary models to base multimodal models (see Fig.~\ref{fig:method}). In general, we utilize standard ReAct framework to let a proprietary model tackle our synthesized multimodal browsing tasks, perform rejection sampling over the generated trajectories, and fine-tune the base multimodal model on the resulting high-quality trajectories. This stage cold-starts the base model with long-horizon, multi-turn tool-use capabilities and prepares it for the subsequent reinforcement learning stage.

\noindent \textbf{ReAct Loop.} In standard ReAct framework, the agent iteratively performs reasoning, action, and observation to solve challenging tasks. In each round, the agent first reasons based on the previous context. It then either calls a tool if it needs additional information or terminates the trajectory by producing a final answer. When a tool call is issued, the agent waits for the observation returned by the tool and continues once the observation is available. In our scenario, we provide 3 real-world web tools, including image search, text search, and visit. For image search, we use the Google Vision Web Detection API on the image in the question; the tool response includes 5 similar image links and corresponding page information that contains those images, including page titles and page links. For text search, we utilize Google Custom Search API; the tool response includes 5 pages related to the query, including the page link, title, and snippet. For visit, the tool returns a summary of the web content based on the provided link and visiting goal describing what to extract from the page. A complete trajectory with $T$ iterations can be defined as:
\begin{align}
\mathcal{H}_T=(\tau_0,a_0,o_0,\dots, \tau_i,a_i,o_i, \dots,\tau_{T},a_{T}),
\end{align}
where $\tau_i$, $a_i$, $o_i$ represent thought, action, and observation in the $i$-th round, respectively. At step $t$, the thought $\tau_t$ and $a_t$ are sampled from a policy based on all previous context, i.e., $\pi(a, t|\mathcal{H}_{t-1})$.

\noindent \textbf{Rejection Sampling.} For the generated ReAct trajectories from the proprietary model, we perform rejection sampling based on answer correctness. Following HLE, we utilize LLM-as-a-Judge to verify whether the final answer is correct, and we keep only trajectories that end with correct answer.

\noindent \textbf{Supervised Fintuning.} Finally, we perform supervised fine-tuning on the base multimodal model with the rejection-sampled full trajectories to instill initial long-horizon tool-use capability. The supervised fine-tuning process minimizes the following objective across all training trajectories:
\begin{equation}
\mathcal{L}_{\text{SFT}} = -\frac{1}{N} \sum_{i=1}^{N} \sum_{t=1}^{T_i} 
\log \pi_{\theta}\big(o_{i,t} \mid o_{i,<t}, q_i \big).
\end{equation}
where $N$ is the number of training trajectories, $o_{i,t}$ denotes the $t$-th token of the output sequence for the $i$-th trajectory, $T_i$ is the total length of the output sequence of the $i$-th training trajectory, $o_{i,<t}$ represents the tokens preceding the $t$-th token of the $i$-th training trajectory, $q_i$ is the input query for the $i$-th training trajectory, and $\pi_{\theta}$ is the model policy parameterized by $\theta$.

\subsection{Reinforcement Learning}

In this stage, we perform reinforcement learning on the fine-tuned multimodal model to further generalize its long-horizon, multi-turn tool-use capabilities. Notably, we conduct RL in real-world web environments, enabling the agent to develop more robust browsing skills and to better solve real-world multimodal user requests. In practice, we adopt the off-the-shelf Group Reward Proximal Optimization (GRPO)~\cite{shao2024deepseekmath} algorithm. Specifically, GRPO performs multiple rollout samplings and optimizes the policy to favor responses with higher assigned rewards, the training objective of GRPO is as follows:

\begin{equation}
\begin{aligned}
J_{\text{GRPO}}(\theta)
=&\;
\mathbb{E}_{q \sim P(Q),\, \{o_i\}_{i=1}^{G} \sim \pi_{\theta_{\text{old}}}(\cdot \mid q)} \\
\Bigg[
\frac{1}{G} \sum_{i=1}^{G} &
\min\Bigg(
\frac{\pi_{\theta}(o_i \mid q)}{\pi_{\theta_{\text{old}}}(o_i \mid q)}\, A_i,\; \\
\operatorname{clip}\!\Big(
\frac{\pi_{\theta}(o_i \mid q)}{\pi_{\theta_{\text{old}}}(o_i \mid q)},\; &
1-\epsilon,\; 1+\epsilon
\Big)\, A_i
\Bigg)
- \beta\, D_{\mathrm{KL}}\!\left(\pi_{\theta}\,\|\,\pi_{\mathrm{ref}}\right)
\Bigg].
\end{aligned}
\end{equation}
\begin{equation}
\mathbb{D}_{KL}\!\left(\pi_{\theta} \,\|\, \pi_{\text{ref}}\right) 
= \frac{\pi_{\text{ref}}(o_i \mid q)}{\pi_{\theta}(o_i \mid q)} 
- \log \frac{\pi_{\text{ref}}(o_i \mid q)}{\pi_{\theta}(o_i \mid q)} - 1,
\end{equation}

where $\pi_\theta$ is the current model policy, $\pi_{\theta_{\text{old}}}$ is the old policy, $G$ is the rollout group size, $q$ is the query, $o_i$ is the $i$-th sampled reponse, $\pi_{\text{ref}}$ is the reference model, $\epsilon$ is the clipping hyper-parameter controlling updating degree, and $\beta$ is the coefficient of Kullback–Leibler (KL) penalty.
\begin{equation}
A_i = \frac{r_i - \text{mean}(\{r_1, r_2, \cdots, r_G\})}
{\text{std}(\{r_1, r_2, \cdots, r_G\})},
\end{equation}

$A_i$ is the normalized advantages computed based on rewards $\{r_1, r_2, \cdots, r_G\}$. 

We conduct strict format checking at each rollout step; trajectories that fail the format checks are terminated immediately. Our reward design considers only final-answer correctness:
\begin{equation}
R_i =  R_{\text{accuracy}}(o_i),
\end{equation}
where $R_{\text{accuracy}}(o_i)$ denotes answer correctness: a trajectory that produces the correct answer receives a reward of 1, and an incorrect answer receives a reward of 0. Following HLE~\cite{phan2025humanity}, we use an LLM-as-a-Judge to evaluate answer correctness.

\subsection{Curation of MM-SearchExam}

To enable more thorough evaluation of multimodal search agents, we curate a challenging multimodal search benchmark, MM-SearchExam. MM-SearchExam is constructed with our proposed Iterative Injection-based Data Synthesis pipeline. Specifically, we perform 10 rounds of iterative extension to make the browsing tasks highly challenging. We also apply data filtering procedure to ensure high quality and difficulty. Additionally, we explicitly remove seeds that appear in our training data to avoid data leakage. In total, MM-SearchExam contains 283 synthesized browsing tasks. Quantitative experiments show that several proprietary models achieve low performance on MM-SearchExam, highlighting difficulty of the benchmark and the effectiveness of our Iterative Injection-based Data Synthesis pipeline in generating highly challenging browsing tasks.

\begin{table*}[!ht]
    \small
    \centering
    \fontsize{8}{9}\selectfont
    \setlength{\tabcolsep}{3mm}
    \caption{
    Comparison between different baselines across 5 multimodal search benchmarks.
    }
    \vspace{-5pt} 
    \begin{tabular}{l|ccccc}
        \toprule
        Model & MMSearch & BrowseComp-VL & MM-BrowseComp & SimpleVQA & MM-SearchExam \\
        \midrule
        \multicolumn{6}{c}{\textit{Open-source Models with Web Tools}} \\
        \midrule
        Qwen3-VL-4B-Thinking & 35.0  & 21.5  & 1.6  & 41.1  & 2.4  \\ 
        Qwen3-VL-8B-Thinking & 39.1  & 24.8  & 5.7  & 45.5  & 4.4  \\ 
        Qwen3-VL-30B-A3B-Thinking & 43.7  & 27.5  & 6.7  & 46.3  & 6.0  \\ 
        Qwen2.5-VL-72B-Instruct & 41.5  & 24.3  & 2.4  & 37.7  & 11.6  \\ 
        InternVL3.5-8B & 29.2  & 17.8  & 1.6  & 38.8  & 3.1  \\ 
        InternVL3.5-38B & 33.3  & 19.3  & 4.9  & 39.7  & 6.3  \\ 
        \midrule
        \multicolumn{6}{c}{\textit{Proprietary Models with Web Tools}} \\
        \midrule
        GPT-4o & 13.4  & 17.3  & 0.8 & 40.0  & 8.1  \\ 
        GPT-5 & 45.2  & 29.0  & 9.3  & 62.3  & 20.8  \\ 
        Gemini-3-Flash & 57.3  & 31.3  & 4.9  & 60.8  & 20.1  \\ 
        Gemini-3-Pro & 59.0  & 35.0  & 10.8  & 64.1  & 23.8  \\ 
        Qwen3-VL-Plus & 50.7  & 28.0  & 6.0  & 58.4  & 18.8  \\ 
        \midrule
        \multicolumn{6}{c}{\textit{Agentic Multimodal Models}} \\
        \midrule
        MMSearch-R1 & 37.5  & 16.0  & 4.9  & 40.0  & 4.5  \\ 
        DeepMMSearch-R1 & - & - & - & 55.8  & - \\ 
        DeepEyesV2 & 40.9  & 22.0  & 8.1  & 44.4  & 4.5  \\ 
        \rowcolor{gray!15} VSearcher (ours) & 47.2  & 30.8  & 6.2  & 46.6 & 19.3 \\ 
        \bottomrule
    \end{tabular}
    \label{tab:main}
\end{table*}

\section{Experiment}

\subsection{Settings}

\noindent \textbf{Implementation Details.}
For rejection sampling fine-tuning, we leverage Gemini-3-Pro-Thinking as the teacher model to generate trajectories for our synthesized browsing tasks due to its strong web-browsing capabilities. For each generated trajectory, we perform rejection sampling based on answer correctness. We obtain 1308 high-quality trajectories for fine-tuning and use LLaMA-Factory to conduct training. For reinforcement learning, we utilize rllm as the training framework. We use a batch size of 32 and perform 8 rollouts per sample. We set the maximum prompt length to 2K and the maximum response length to 28K. The maximum number of steps in the ReAct loop is set to 30, and the maximum trajectory time is set to 2 hours 30 minutes. The learning rate is set to 2e-6. The KL-penalty coefficient is set to 0.001. We use a rollout temperature of 0.6 and top\_p of 0.9. We use Qwen2.5-72B-Instruct as the judge model for verifying answer correctness. We utilize Qwen3-VL-8B-Thinking as our base models. We use 16 H100 GPUs for all experiments.

\noindent \textbf{Tool Implementation}
For text search, we use the Google Custom Search API as the backend. For each query, we return 5 relevant search results, including the page link, page title, and page snippet which provides a brief summary of the page. For image search, we use the Google Vision Web Detection API. For each input image, we return the 5 most similar image URLs and the corresponding pages that contain these images, including their page links and page titles. For the visit tool, the inputs are a webpage URL and a specific visit goal. We first call the JINA API on the URL, which parses the webpage content and returns it in clean markdown format. We then use Qwen2.5-72B-Instruct as a summarization model to extract content relevant to the visit goal. The resulting summary is returned as the final response of the visit tool.

\noindent \textbf{Baselines} MMSearch-R1~\cite{wu2025mmsearch} is a pioneering work that utilizes RL to train multimodal search agent. It is equipped with image search and text search. Specifically, it mainly focuses on a two-step tool-use pattern: first performing image search, followed by text search. DeepMMSearch-R1~\cite{narayan2025deepmmsearch} is a follow-up work to MMSearch-R1. It primarily improves image-search effectiveness by incorporating an image-cropping tool and enabling self-reflection and self-correction during rollouts. It emphasizes multi-turn tool-use capabilities. The agent is equipped with image search, text search, and image cropping tool. DeepEyesV2~\cite{hong2025deepeyesv2} is a comprehensive study on building multimodal agentic models. It utilizes Python code to perform image operations such as cropping, and it is also equipped with image search and text search.

\subsection{Main Results}

We present comparison with various baselines in Tab.~\ref{tab:main}. We compare VSearcher against open-source multimodal models equipped with web tools, proprietary models equipped with web tools, and recent agentic multimodal models across 5 benchmarks. All agentic models share the same tool implementation to ensure fair comparison. We observe that VSearcher outperforms strong agentic multimodal models by clear margin across various benchmarks. Notably, VSearcher also surpasses strong proprietary models, \eg, GPT-5, on several benchmarks, including MMSearch and BrowseComp-VL. These results demonstrate the strong multimodal browsing capabilities of VSearcher. Moreover, the strong performance of VSearcher validates the effectiveness of rejection-sampling fine-tuning with high-quality teacher trajectories for instilling initial multi-turn tool-use abilities, as well as reinforcement learning for further enhancing multi-turn multimodal browsing skills.

\section{Detailed Analysis}

\begin{figure}
    \centering
    \includegraphics[width=\linewidth]{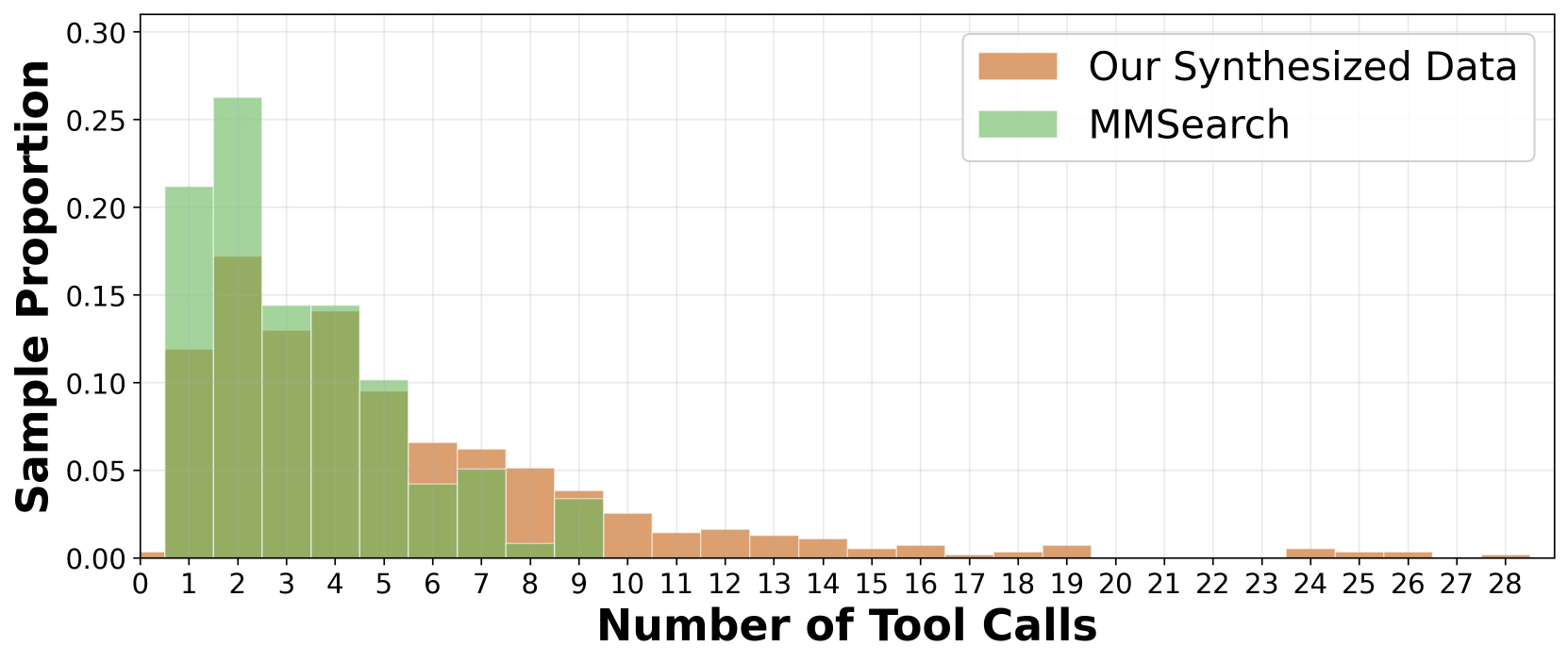}
    \vspace{-10pt}
    \caption{
    Tool call distribution of Gemini-3-Pro-Thinking in solving our synthesized data and MMSearch questions.
    }
    \label{fig:tool_call_number}
    \vspace{-10pt}
\end{figure}

\noindent \textbf{Synthesized Data Quality.} We assess the quality of our synthesized data by comparing the number of tool calls required to solve each browsing task (see Fig.~\ref{fig:tool_call_number}). We perform this analysis by utilizing Gemini-3-Pro-Thinking to solve both our synthesized data and MMSearch~\cite{jiang2024mmsearch} under the ReAct framework. We observe that (1) the distribution of tool-call counts for our synthesized data is closely calibrated to that of MMSearch. This strong alignment between the complexity of our agent training data and the target evaluation benchmark helps build appropriate multi-turn multimodal browsing skills for real-world tasks. It also indicates that our synthesis pipeline can generate benchmark-level difficulty data at scale in a fully automatic manner. (2) The tool-call distribution for our synthesized data exhibits a long-tail pattern: some synthesized tasks require Gemini-3-Pro-Thinking to utilize more than 20 tool calls to solve. These highly challenging samples can help VSearcher develop more complex multi-turn browsing skills during rejection-sampling fine-tuning and RL.

\begin{figure}
    \centering
    \includegraphics[width=\linewidth]{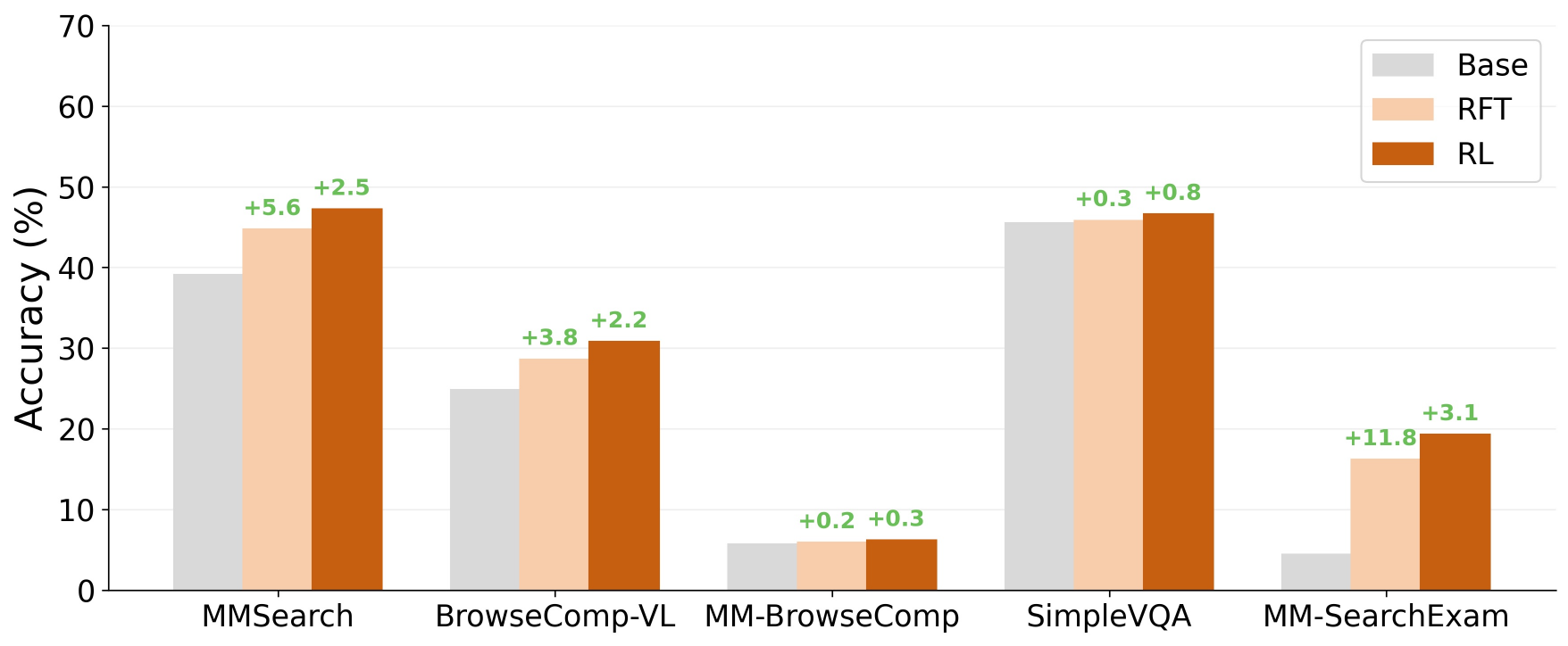}
    \vspace{-10pt}
    \caption{
    Performance comparison between base model, RFT model and RL model.
    }
    \label{fig:base_rft_rl_accuracy}
    \vspace{-15pt}
\end{figure}

\noindent \textbf{Performance Improvement Analysis.} We analyze accuracy improvements across three stages: the base model, the model after rejection-sampling fine-tuning (RFT), and the model after RL (see Fig.~\ref{fig:base_rft_rl_accuracy}). We observe a progressive improvement from the base model to RFT and then to RL. After RFT, the model consistently achieves clear gains over the vanilla base model across diverse benchmarks. Since our RFT data are synthesized by the Iterative Injection-based Data Synthesis pipeline, the high complexity of the synthesized browsing tasks provides a strong foundation for improving browsing skills. Moreover, the teacher model, Gemini-3-Pro-Thinking, transfers its strong browsing behaviors to the base model, which helps the base model learn to solve complex browsing tasks. After RL, model obtains further improvements in answer accuracy. This suggests that RL further generalizes the initial multi-turn browsing skills learned during RFT. Interacting with the real-world web environment also strengthens adaptive tool selection and overall browsing capability.

\begin{figure}
    \centering
    \includegraphics[width=\linewidth]{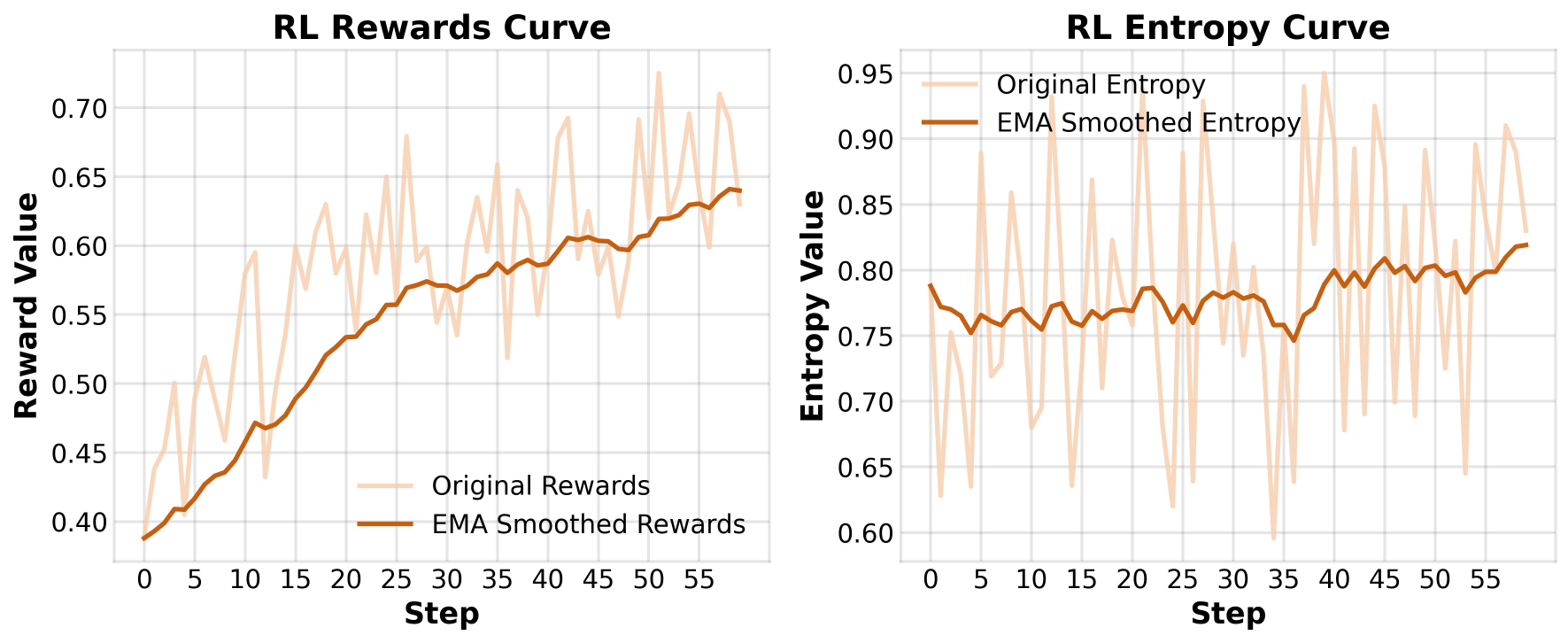}
    \vspace{-10pt}
    \caption{
    Our RL training dynamic, including the reward curve and RL entropy.
    }
    \label{fig:rl_reward_entropy_combined}
    \vspace{-10pt}
\end{figure}

\noindent \textbf{RL Training Dynamics.} We present the reward curve and RL entropy during training (see Fig.~\ref{fig:rl_reward_entropy_combined}). We observe that the reward exhibits a steady increasing trend throughout RL, indicating that reinforcement learning in a real-world web environment effectively improves the multi-turn browsing skills of our multimodal agent. Since RL training utilizes only our synthesized data, these results also reflect the high complexity and quality of the synthesized data, which substantially contribute to stronger agentic browsing capabilities. Regarding entropy, it remains stable over the entire RL process, suggesting that our training is stable. The agent maintains a reasonable level of exploration, avoiding both entropy collapse and overly aggressive exploration.

\begin{figure*}
    \centering
    \includegraphics[width=0.9\linewidth]{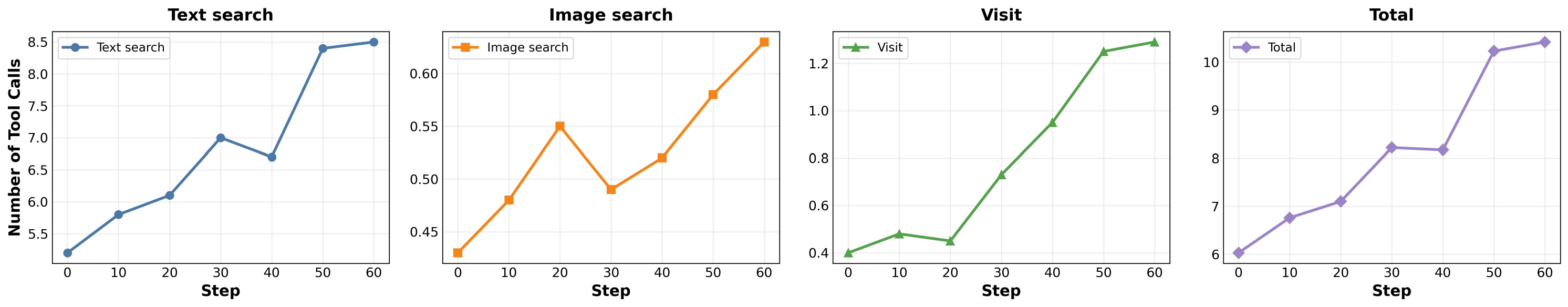}
    \vspace{-5pt}
    
    \caption{
    Analysis of tool call count for different tools during the RL process.
    }
    \vspace{-10pt}
    \label{fig:tool_calls_by_type_vs_step}
\end{figure*}

\noindent \textbf{Tool Call Number Analysis.} We analyze tool-use dynamics during RL (see Fig.~\ref{fig:tool_calls_by_type_vs_step}). We observe that most tool calls are text searches. This is because text search remains the primary source of information during information seeking: it is critical for obtaining additional cues after image search and for progressively solving complex browsing tasks. The average number of image-search calls is around 0.6. Since we only conduct image search for image provided in the question, agent usually perform at most one time of image search for one provide question. In contrast, number of visit calls increases steadily over training, indicating that model increasingly learns to leverage visit tool to gather more detailed evidence and to verify intermediate conclusions.

\section{Related Work}

\noindent \textbf{Deep Research Agent.}
The construction of deep research agents can be categorized as static or dynamic~\cite{huang2025deep,xu2025comprehensive,zhang2025deep} Static deep research agents follow explicitly predefined pipelines~\cite{arslan2024survey,cuconasu2024power,gao2023retrieval}, which makes them more suitable for specific, structured scenarios, such as scientific research. For example, AI Scientist~\cite{lu2024ai} treats scientific discovery as a fixed sequence of steps, including ideation, experimentation, and reporting. Dynamic workflows focus on adaptive task planning: they dynamically design and schedule the solution pipeline based on the task, user feedback, and evolving context. Planning-only methods directly generate task plans from the initial user query, which is used by most current deep research systems, such as Grok DR, H2O, and Manus. The intent-to-planning strategy first clarifies user intent through targeted questions and then generates a task plan based on refined requirements, which is utilized by OpenAI DR. Unified intent-planning combines these two strategies by first generating a preliminary plan and then engaging user to revise it, which is adopted by Gemini DR. Different from existing works, our method focuses on the multimodal domain, solving complex multimodal questions by dynamically using various tools, such as image search, text search, and visit.

\noindent \textbf{Multimodal QA Benchmark}
Most existing multimodal QA benchmarks focus on evaluating single-step prediction or limited retrieval ability, with little emphasis on long-horizon reasoning and planning~\cite{li2024survey,li2024survey}. Benchmarks such as OK-VQA~\cite{marino2019ok} and A-OKVQA~\cite{schwenk2022okvqa} emphasize evaluation of static model knowledge and direct answer prediction, without requiring complex retrieval or reasoning. Recent efforts have begun to shift attention toward model reasoning and tool-use capabilities. MMT-Bench~\cite{ying2024mmt} covers large-scale, planning-oriented tasks across multiple domains; however, its multiple-choice output format restricts thorough evaluation of long-horizon reasoning. Dyn-VQA~\cite{li2024benchmarking} constructs adaptive query tasks but remains limited in both multimodal diversity and scale. While benchmarks such as MMMU~\cite{yue2024mmmu} and MMMU-Pro~\cite{yue2025mmmu} further reveal limitations of recent MLLMs~\cite{chen2024internvl} through challenging, domain-specific tasks, no existing benchmark simultaneously supports multi-step reasoning, large-scale coverage, and fully automated curation with rigorous quality control. In contrast to existing benchmarks, we propose a challenging benchmark designed to evaluate long-horizon reasoning and tool-use capabilities of MLLMs, which is large-scale and constructed in fully automated manner.

\noindent \textbf{Large Vision-Language Model}
Large Vision-Language Models (LVLMs), which are built upon Large Language Models (LLMs), have demonstrated strong general performance in multimodal understanding~\cite{zhang2024vision,liu2025survey,zhang2024vl,zhang2025uncertainty}. Current LVLMs achieve high accuracy across a wide range of vision tasks, including general visual capabilities, visual question answering, OCR, grounding, and visual reasoning. Despite these advances, such models typically rely on static knowledge and are unable to incorporate up-to-date information. Consequently, recent research has focused on equipping LVLMs with tool-use capabilities, transforming them into autonomous agents that can dynamically acquire external knowledge through tool usage. In contrast to existing approaches, our method enhances base LVLMs with both text search and image search tools in a real-world web environment and facilitates long-horizon, multi-turn tool usage, significantly improving the ability of model to solve complex multimodal problems.

\section{Conclusion}

In this paper, we propose VSearcher, a long-horizon, multi-turn, tool-using multimodal search agent trained with reinforcement learning in real-world web environment. VSearcher can adaptively leverage text search, image search, and visit tools to retrieve web information and solve challenging tasks. We propose a systematic post-training framework to convert static multimodal model into multimodal search agent. In particular, we propose Iterative Injection-based Data Synthesis pipeline, together with rigorous data filtering strategies, to automatically synthesize highly challenging and high-quality multimodal browsing tasks at scale. We further adopt rejection-sampling fine-tuning to transfer strong multi-turn browsing ability from proprietary teacher model to base multimodal models, and apply reinforcement learning to further generalize the tool-use capabilities. For more thorough evaluation of multimodal search agent, we curate a highly challenging benchmark, MM-SearchExam, on which several recent proprietary models yield low performance. Across extensive evaluations on multiple multimodal search benchmarks, VSearcher consistently outperforms recent multimodal search agents and even surpasses several proprietary models.

\clearpage

\section*{Impact Statement}

We propose VSearcher, a multimodal search agent with long-horizon, multi-turn tool-use capabilities for solving complex information-seeking tasks. Our work focuses on utilizing rejection-sampling fine-tuning and reinforcement learning to build multi-turn tool-use abilities in multimodal models, and we do not foresee significant ethical concerns arising from our method.






\bibliography{example_paper}
\bibliographystyle{icml2026}

\newpage
\appendix
\onecolumn



\section{Implementation Detail}

\subsection{Tool Implementation}

\noindent \textbf{Image Search.} We utilize Google Vision Web Detection API to implement our image search tool. The input to this service is the URL of the query image, and it returns web pages that contain images similar to the input. We then post-process the returned pages. Specifically, we iterate over the pages and, for each page, verify whether the similar images it contains are reachable from our server network and correspond to valid image files. If so, we add the image and its source page to the returned list. Specifically, we only select one image for one returned page. For each input image, we collect information of five similar web images, including the image URL, the URL of the page containing the image, and the page title.

\noindent \textbf{Text Search.} Our text search tool is implemented based on Google Custom Search API. The text search tool input is the text query to be searched. For each query, the tool returns the top 5 most relevant web pages. Each result includes the page URL, title, and a snippet that provides a concise preview of the content. The Google Custom Search API has a daily rate limit of 10{,}000 requests; we use multiple API keys to ensure sufficient daily budget for RL training. The API is provided through Google Cloud, which offers a \$300 free quota and is convenient for initial usage.

\noindent \textbf{Visit.} We leverage the Jina API to implement our visit tool. The input to the visit tool is a target URL and a specific goal for visiting the page. We first query the Jina API with the URL, which returns the cleaned full content of the web page in Markdown format. Because the returned content can be extremely long, directly leveraging the full content as the tool response in the ReAct process can quickly exhaust the model context limit. Therefore, we further utilize an LLM, Qwen2.5-72B-Instruct, as a summarization model: it takes the full web content and the visit goal as input and summarizes the relevant information conditioned on the goal. This summary is returned as the final tool response.

\subsection{Iterative Injection-based Data Synthesis}

\noindent \textbf{Seed Selection.} We obtain rare entities from Wikidata to serve as seeds, which helps synthesize challenging multimodal browsing tasks. For an entity in Wikidata, `sitelink' refers to the number of Wikipedia page links in different languages, and `statement' refers to the number of atomic facts associated with the entity. We control entity rarity by requiring the number of sitelinks to be below a threshold, while ensuring sufficient coverage by requiring the number of statements to be above a threshold. In practice, we select seed entities with sitelink $\leq 10$ and statement $\geq 20$. We utilize SPARQL queries to access the official Wikidata Query Service and retrieve seed entities.

\noindent \textbf{Initial Question Generation.} For each seed entity, we retrieve its Wikipedia page content and generate a simple QA pair based on the content, which serves as the starting point for subsequent iterative extensions. Specifically, we download an offline Wikipedia dump (2025-08) and serve Wikipedia locally for more convenient content acquisition. After obtaining the Wikipedia content for the seed entity, we utilize Qwen2.5-72B-Instruct to generate a simple QA pair about the entity.

\noindent \textbf{Text Information Injection.} Starting from the initial QA pair, we perform several rounds of text information injection to transform it into a complex browsing task. Each round contains three steps: entity selection, information parsing, and question transformation. In each round, we first utilize Qwen2.5-72B-Instruct to select an entity from the current question. We provide examples in the prompt to encourage selecting meaningful entities (\eg, World War II) rather than ambiguous ones (\eg, theory). After selecting an entity, we retrieve its Wikipedia content from the offline index and parse a specific fact about the entity with Qwen2.5-72B-Instruct. For information parsing, we explicitly prompt the LLM to extract less well-known information, which increases the challenge of the extended question and improves the difficulty of the final browsing task. We also encourage the extracted information to include as many entities as possible, so that subsequent injection rounds have additional entities to extend. We also prompt LLM to extract information that is concise and in one sentence, to avoid over-long question. Finally, we prompt Qwen2.5-72B-Instruct to transform the current question by hiding the selected entity and replacing it with the parsed information.

\noindent \textbf{Image Injection.} After the text information injection process, we perform image injection to transform the question into a multimodal browsing task. We first prompt Qwen2.5-72B-Instruct to select an image entity from the question. Notably, we explicitly prompt the LLM to choose an entity that is critical for answering the question. We then prompt the LLM to replace the selected entity with a phrase such as `shown in the image', and retrieve Wikipedia image of the selected entity as the image in the multimodal browsing task. This design ensures that the injected image is critical for answering the question correctly, which faciliates multimodal search agent to focus on the multimodal input.

\noindent \textbf{Difficulty Level Design.} We explicitly design 3 difficulty levels for the synthesized data to ensure that our trained search agent can handle tasks requiring different degrees of browsing effort. Specifically, we construct 3 levels of multimodal browsing tasks: Easy, Medium, and Hard. For Easy questions, we perform one round of text information injection. For Medium questions, we perform three rounds. For Hard questions, we perform five rounds.

\noindent \textbf{Data Filtering.} We design 4 filtering criteria to ensure the difficulty and quality of the synthesized data. First, we feed the synthesized question to Qwen2.5-VL-72B-Instruct and prompt it to answer directly; we filter out questions that it can answer correctly without tool use. These questions typically require few tool calls and are ineffective for improving multi-turn tool-use capabilities. Second, we input only the text part of the synthesized question to Qwen2.5-72B-Instruct and filter out questions that can be answered correctly from text alone. Third, we filter questions with overly simple images (\eg, images that contain only easily recognizable text). Specifically, we utilize Qwen2.5-VL-72B-Instruct as an image judge and filter samples that it deems too simple. Finally, we filter corner cases where the answer is explicitly revealed in the question, which can happen when text information injection injects the answer into the question.

\subsection{Rejection Sampling Finetuning}

\noindent \textbf{Teacher Model} We utilize Gemini-3-Pro-Thinking as the teacher model to generate multi-turn tool-calling trajectories. We choose it for two reasons. (1) We observe strong multi-turn browsing capabilities in Gemini-3-Pro-Thinking, including adaptive image search for complex image inputs, dynamically adjusting text search queries in creative ways to retrieve useful information, and issuing appropriate visit actions to obtain more concrete evidence. (2) We find that it achieves strong performance on various browsing-focused benchmarks, surpassing other proprietary models by clear margin. These advantages make it well-suited as a teacher model for generating high-quality information-seeking trajectories.

\noindent \textbf{Trajectory Generation} We utilize a standard ReAct process with Gemini-3-Pro-Thinking to generate information-seeking trajectories. We set the max number of tool calls to 30. The trajectory timeout is set to 2 hours 30 minutes. The max context length is set to 200K. We set the rollout temperature to 0.6, top\_p to 0.95, the max new tokens to 10{,}000, and the presence penalty to 1.5 to reduce repetition. Notably, we enable strict format checking during each round of rollout process. For format checking, each response must contain exactly one pair of <think> and </think> tags. After the thinking section, the response must contain either <tool\_call></tool\_call> or <answer></answer>; having both is not allowed. All XML tags must be properly closed. The tag order must be either <think>...</think><tool\_call>...</tool\_call> or <think>...</think><answer>...</answer>. Each response must contain exactly one tool call. If a tool call is present, its content must be a valid JSON object. If a model response fails format checking, we terminate the rollout immediately. This strict format control ensures that the cold-started model learns the correct response format.

\noindent \textbf{Rejection Sampling} We perform rejection sampling on the generated trajectories to retain high-quality trajectories as effective supervision signals for the base models during finetuning process. Rejection sampling is based on the final answer correctness. We evaluate answer correctness with an LLM-as-a-judge approach, following the practice in BrowseComp. We utilize Qwen2.5-72B-Instruct as the judge model. We retain only trajectories that yield correct final answer.

\noindent \textbf{Supervised Finetuning} After rejection sampling, we collect 1{,}301 high-quality trajectories from Gemini-3-Pro-Thinking. We then perform supervised fine-tuning on the base models with these trajectories to instill initial multi-turn tool-calling abilities. We utilize LLaMA-Factory as the SFT framework. Notably, we perform full fine-tuning to improve instruction-following ability. We set the per-device batch size to 1 and the gradient accumulation steps to 2. The learning rate is set to 1e-5. The number of epochs is set to 5. We utilize a warmup ratio of 0.1. The cutoff length is set to 20{,}480.

\subsection{RL Training Details}

We utilize rllm as the RL training framework. We choose this framework because it is specifically designed for agentic RL training. In the latest version of rllm, it is convenient to convert a multi-turn tool using ReAct process into multi-turn tool-use RL training.

For RL, the training batch size is set to 32, and number of rollouts is 8 per training sample. The max prompt length is set to 2K, and the max new token is set to 28K, resulting in max model length of 30K. The max number of tool calls is set to 30. The total trajectory timeout is set to 2 hours 30 minutes, since the high degree of parallelism during RL training can slow down the rollout process, especially for trajectories with extensive tool calls. The learning rate is set to 2e-6. We disable the KL penalty to encourage more exploration during RL. The rollout temperature is set to 0.6. We filter out rollout trajectories that exceed the max prompt length, exceed the max response length, exceed the max tool-call limit, time out, or encounter unexpected errors. These trajectories are utilized only for advantage computation and are excluded from the loss calculation. Finally, we use 1K synthesized samples from our proposed Iterative Injection-based Data Synthesis for RL training, with Easy:Medium:Hard ratio of 4:3:3. We utilize Qwen3-VL-2B-Thinking, Qwen3-VL-4B-Thinking, and Qwen3-VL-8B-Thinking after rejection-sampling fine-tuning as the RL base models. We utilize 16 H100s (80G) for all experiments.

\subsection{Benchmark}

\noindent \textbf{MMSearch.} MMSearch is a pioneering benchmark for evaluating multimodal browsing capabilities. It contains 300 QA questions spanning 14 topics, such as finance and entertainment. It includes 171 samples with image input and 129 samples without image. Since we focus on multimodal search, we evaluate on the 171 samples with image input.

\noindent \textbf{BrowseComp-VL.} BrowseComp-VL is a recently proposed and challenging multimodal search benchmark. It contains two subsets, Level1 and Level2. The Level1 subset includes 199 samples that already require challenging multi-hop reasoning. The Level2 subset includes 200 samples and is more difficult than Level1 because key information is further fuzzed to increase question difficulty. We utilize all samples in this benchmark for evaluation since they all include image input.

\noindent \textbf{MM-BrowseComp.} MM-BrowseComp contains 224 challenging, hand-crafted questions designed to evaluate multimodal retrieval ability. The questions are constructed in order to make sure text alone is insufficient to answer correctly. We filter out samples without image input, resulting in 122 samples for evaluation.

\noindent \textbf{SimpleVQA.} SimpleVQA is designed to evaluate the factual question-answering ability of multimodal models. It consists of questions with short natural-language descriptions of factual knowledge. It contains 2{,}025 samples spanning 9 topics and is designed to have short, concrete answers that are easy to verify. We randomly sample 180 questions from its English subset for evaluation.

\section{Prompt}

This is the system prompt for our multimodal search agent, used for both the ReAct inference and RL training processes.

\begin{tcolorbox}[breakable,title=System Prompt]

You are a multimodal deep research agent. Given a user question, you should conduct thorough searches across various information sources on the real-world internet, perform analysis and reasoning, and give accurate answers to the user question.

\vspace{1em}

Workflow:

- You should first conduct reasoning within $<$think$>$$</$think$>$ tags. This includes analysis of the given question, interpretation of tool-returned information, and analysis of what actions need to be taken next.

- If you think you need to call a tool to provide you with additional information, you should call the tool within $<$tool\_call$>$$</$tool\_call$>$ tags.

- The returned information from tools will be returned to you within $<$tool\_response$>$$</$tool\_response$>$ tags.

- If you think you have gathered enough information and are confident you can answer the question, provide your final answer within $<$answer$>$$</$answer$>$ tags (e.g., $<$answer$>$Titanic$</$answer$>$). Do not provide explanations in $<$answer$>$$</$answer$>$ tags.

\vspace{1em}

Guidelines:

- If you need to call a tool, you should call only one tool at a time. Do not call multiple tools at the same time.

- You should not provide tool responses yourself. You should only call tools and wait for the tool response.

\vspace{1em}

Tool set:

$<$tools$>$

\{``type'': ``function'', ``function'': \{``name'': ``text\_search'', ``description'': ``Perform a Google web search and return a string of the top search results.'', ``parameters'': \{``type'': ``object'', ``properties'': \{``query'': \{``type'': ``string'', ``description'': ``The search query.''\}\}, ``required'': [``query'']\}\}\}

\{``type'': ``function'', ``function'': \{``name'': ``image\_search'', ``description'': ``Perform a Google image search on the given image and return a list of URLs of similar images, along with the titles and links of the pages where they appear. Note that the image search is only conducted on the initial image provided by the user, so no parameters are needed for this tool.''\}\}

\{``type'': ``function'', ``function'': \{``name'': ``visit'', ``description'': ``Visit a webpage and return a summary of its content.'', ``parameters'': \{``type'': ``object'', ``properties'': \{``url'': \{``type'': ``string'', ``description'': ``The URL of the webpage to visit.''\}, ``goal'': \{``type": ``string'', ``description'': ``The specific information goal for visiting the webpage.''\}\}, ``required'': [``url", ``goal'']\}\}\}

$</$tools$>$

\vspace{1em}

For each function call, return a JSON object with the function name and arguments inside $<$tool\_call$>$$<$/tool\_call$>$ tags:

$<$tool\_call$>$

\{``name'': $<$function-name$>$, ``arguments'': $<$args-json-object$>$\}

$<$/tool\_call$>$

\vspace{1em}

Current date:

\end{tcolorbox}

This is the LLM-as-Judge prompt we adopt, following the prompt in HLE and BrowseComp.

\begin{tcolorbox}[breakable,title=LLM-as-Judge Prompt]

Judge whether the following $[$response$]$ to $[$question$]$ is correct or not based on the precise and unambiguous $[$correct\_answer$]$ below.
\vspace{1em}

$[$question$]$: \{question\}
\vspace{1em}

$[$response$]$: \{response\}
\vspace{1em}

Your judgement must be in the format and criteria specified below:
\vspace{1em}

extracted\_final\_answer: The final exact answer extracted from the $[$response$]$. Put the extracted answer as `None' if there is no exact, final answer to extract from the response.
\vspace{1em}

$[$correct\_answer$]$: \{correct\_answer\}
\vspace{1em}

reasoning: Explain why the extracted\_final\_answer is correct or incorrect based on $[$correct\_answer$]$, focusing only on if there are meaningful differences between $[$correct\_answer$]$ and the extracted\_final\_answer. Do not comment on any background to the problem, do not attempt to solve the problem, do not argue for any answer different than $[$correct\_answer$]$, focus only on whether the answers match.
\vspace{1em}

correct: Answer `yes' if extracted\_final\_answer matches the $[$correct\_answer$]$ given above, or is within a small margin of error for numerical problems. Answer `no' otherwise, i.e. if there if there is any inconsistency, ambiguity, non-equivalency, or if the extracted answer is incorrect.
\vspace{1em}

confidence: The extracted confidence score between 0\% and 100\% from $[$response$]$. Put 100 if there is no confidence score available.

\end{tcolorbox}

This is the prompt for the summarization model to extract relevant information from the full web page content.

\begin{tcolorbox}[breakable,title=Summary Prompt in Visit Tool]

You are an expert in summarizing web page information related to a user goal.
\vspace{1em}

Given the following web page content and a specific user goal for visiting this page, you should carefully read the web page content, collect relevant details with respect to the user goal, and generate a paragraph-level summary of those details in response to the user goal. The generated summary should be accurate, clear, and logically coherent.
\vspace{1em}

Web page content:

[WEB PAGE CONTENT]
\vspace{1em}

User goal:

[USER GOAL]

\end{tcolorbox}

This is the prompt used to generate the initial QA pair for seed entity based on its Wikipedia content.

\begin{tcolorbox}[breakable,title=Prompt for Initial QA Pair Parsing]

Generate a question and answer pair based on the page content and the provided entity.
\vspace{1em}

Guidelines:

- The question should be about the provided entity.

- Question should be one sentence.

- The question should be specific so that there exist only one answer.

- The answer should be a word or a short phrase. which is easy to verify correctness.

- The answer should be concise and specific.

- Return the question and answer pair in json format.
\vspace{1em}

The json format should be like:

\{

    ``question'': ...,
    
    ``answer'': ...
    
\}

Do not provide any other information.

\vspace{1em}

Page content:

[PAGE CONTENT]
\vspace{1em}

Entity:

[ENTITY]

\end{tcolorbox}

This is the prompt for selecting entity for text information injection.

\begin{tcolorbox}[breakable,title=Prompt for Entity Selection]

Select a entity from the given question.
\vspace{1em}

Guidelines:

- Only output the entity, no need for other explantion.

- The entity should be substring of the given question, do not answer the question.
\vspace{1em}

Example:

Ukraine, World War II, Ivy League, PageRank, Sverker Johansson, Virginia Woolf, Reykjavík, Borobudur
\vspace{1em}

Do not select entity like this:

German-born theoretical physicist, Polish-born, theory, Roman emperor from 98 to 117 AD, 
\vspace{1em}

Text:

[TEXT]

\end{tcolorbox}

This is the prompt for parsing specific information of selected entity from its Wikipedia content.

\begin{tcolorbox}[breakable,title=Prompt for Entity Information Parsing]

Parse information of the entity from the given text.
\vspace{1em}

Guidelines:

- The information should be rarely known.

- The information should contain more entities as possible.

- The information should be short and concise, in one sentence.

- Only output the information, no need for other explantion.
\vspace{1em}

Parse information like this:

First-lap contact with Lewis Hamilton dropped Lando Norris to the back at the Spanish Grand Prix 2023.

A member of the McLaren Young Driver Programme since 2017, Lando Norris joined McLaren in 2019 to partner Carlos Sainz Jr..
\vspace{1em}

Do not parse information like this:

Vladimir Putin has served as President of Russia since 2012.

As observational evidence for a dynamic universe was lacking at the time, Einstein introduced a new term.

Following the discovery of the recession of the galaxies by Edwin Hubble in 1929, Einstein abandoned his static model of the universe, and proposed two dynamic models of the cosmos, the Friedmann–Einstein universe of 1931  and the Einstein–de Sitter universe of 1932. In each of these models, Einstein discarded the cosmological constant, claiming that it was "in any case theoretically unsatisfactory".
\vspace{1em}

Text:

[TEXT]
\vspace{1em}

Entity:

[ENTITY]

\end{tcolorbox}

This is the prompt for injecting the parsed information into original question and hiding the selected entity.

\begin{tcolorbox}[breakable,title=Prompt for Text Information Injection]

Transform the question based on the entity and its information.
\vspace{1em}

Guidelines:

- The entity should be hidden from the question.

- Replace the entity with its information.

- Only output the transformed question, no need for other explantion.
\vspace{1em}

Use the part of the information that is less known.

Example:

Full Information:

Founded by Jimmy Wales and Larry Sanger in 2001, Wikipedia began as a complementary project for Nupedia, switching from its own license to the GNU Free Documentation License at the urging of Richard Stallman.

Use information like this:

switching from its own license to the GNU Free Documentation License at the urging of Richard Stallman.

Do not use information like this:

Founded by Jimmy Wales and Larry Sanger in 2001
\vspace{1em}

Use the part of the information that contains more entities.

Example:

Full Information:

The Mojave phone booth, originally set up in 1948 for volcanic cinder miners, became an internet sensation in 1997 after being featured in a New York Times article, leading to its removal by Pacific Bell in 2000 due to environmental concerns.

Use information like this:

became an internet sensation in 1997 after being featured in a New York Times article, leading to its removal by Pacific Bell in 2000 due to environmental concerns.

Do not use information like this:

The Mojave phone booth, originally set up in 1948 for volcanic cinder miners
\vspace{1em}

Question:

[QUESTION]
\vspace{1em}

Entity:

[ENTITY]
\vspace{1em}

Information:

[INFORMATION]

\end{tcolorbox}

This is the prompt for selecting image entity from extended question.

\begin{tcolorbox}[breakable,title=Prompt for Image Entity Selection]

Select a entity from the given question.
\vspace{1em}

Guidelines:

- Only output the entity, no need for other explantion.

- The entity should be substring of the given question, do not answer the question.

- Select the entity that is more critical to the correct answering of the question.
\vspace{1em}

Example:

Ukraine, World War II, Ivy League, PageRank, Sverker Johansson, Virginia Woolf, Reykjavík, Borobudur

\vspace{1em}

Do not select entity like this:

German-born theoretical physicist, Polish-born, theory, Roman emperor from 98 to 117 AD, 
\vspace{1em}

Text:

[TEXT]

\end{tcolorbox}

This is the prompt for hiding image entity and replacing it with phase like `shown in the image'.

\begin{tcolorbox}[breakable,title=Prompt for Image Injection]

Transform the question based on the entity.
\vspace{1em}

Guidelines:

- The entity should be hidden from the question.

- Replace the entity with phase like `shown in the image'.

- The question should be rephrased to keep logically coherent and easy to understand.
\vspace{1em}

Question:

[QUESTION]
\vspace{1em}

Entity:

[ENTITY]

\end{tcolorbox}

This is the prompt for obtaining direct answer from model.

\begin{tcolorbox}[breakable,title=Prompt for Direct Answer]

[QUESTION]
\vspace{1em}

Please directly answer this question. Answer with a short phrase. Do not provide any other information.

\end{tcolorbox}

This is the prompt for evaluating the image complexity during data filtering process.

\begin{tcolorbox}[breakable,title=Prompt for Image Evaluation]

I am currently building a training dataset for a multimodal deep-research agent. I want the images in the training samples to be complex so that they encourage the agent to perform image search to obtain more information. I want to filter out samples with images that are too simple.
\vspace{1em}

Now, evaluate the given image.
\vspace{1em}

Answer ``yes'' if the image is simple (e.g., it is the flag of the United States or contains only text that is easy to parse).
Answer ``no'' if the image is complex, information-rich, or hard to interpret.
\vspace{1em}

Just provide the final answer, do not provide any other information.

\end{tcolorbox}

\section{Synthesised Data}

\begin{tcolorbox}[breakable,title=Case for Easy Data]

\textbf{Question:} In which county is the state park located that was the first and largest park acquired under the Project 70 Land Acquisition and Borrowing Act, featuring the river shown in the image, which is the busiest whitewater east of the Mississippi River?

\textbf{Image:}

\begin{center}
\includegraphics[width=0.4\linewidth]{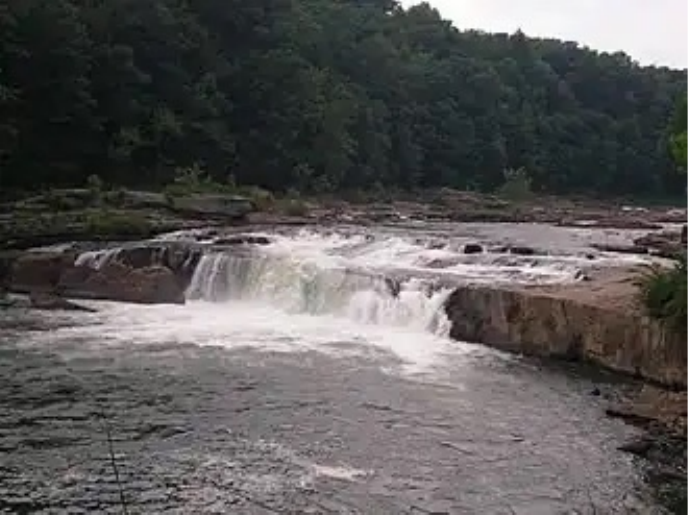}
\end{center}

\textbf{Answer:} Fayette County

\end{tcolorbox}

\begin{tcolorbox}[breakable,title=Case for Medium Data]

\textbf{Question:} What is the common name of the grass that was described by explorers during the expedition commissioned by the third President of the United States, who served from 1801 to 1809 and orchestrated the Louisiana Purchase, which included the Corps of Discovery led by Captain Meriwether Lewis and Second Lieutenant William Clark, as shown in the image?

\textbf{Image:}

\begin{center}
\includegraphics[width=0.4\linewidth]{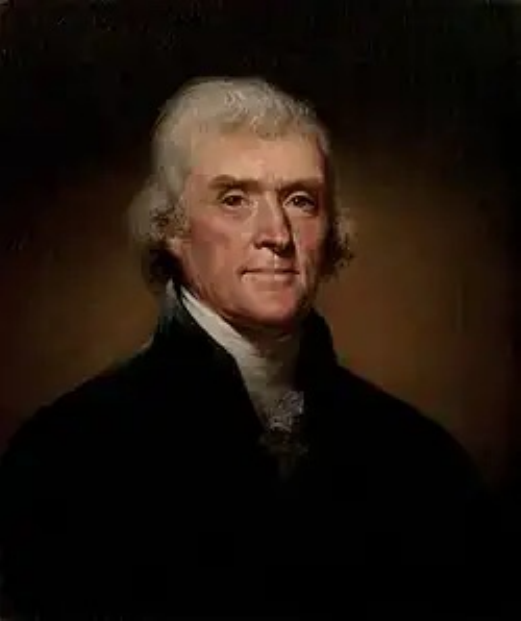}
\end{center}

\textbf{Answer:} needle-and-thread grass

\end{tcolorbox}

\begin{tcolorbox}[breakable,title=Case for Medium Data]

\textbf{Question:} In which year was the moth species, similar to the one described by the son of a British botanist and entomologist who was the first to describe the insect order Thysanura in 1840, and Mary, daughter of Edward Betts, in 1847, but with a deeper yellow underside and less conspicuous wing markings, described by the British entomologist who compiled the major work \"Lepidoptera Indica\" on butterflies from the region shown in the image, with many plates produced by his son, E. C. Knight, and John Nugent Fitch?

\textbf{Image:}

\begin{center}
\includegraphics[width=0.4\linewidth]{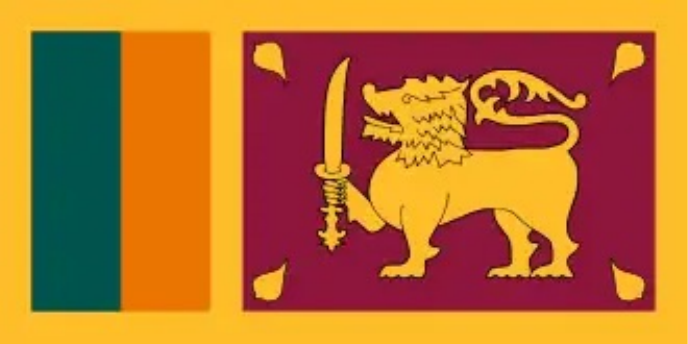}
\end{center}

\textbf{Answer:} 1879

\end{tcolorbox}

\section{Case Study}

\begin{tcolorbox}[breakable,title=Case Trajectory in MMSearch]

\textcolor{black}{\textbf{Question:} Are the art work in the image and Marcel Duchamp's Bicycle Wheel both belong to installation art?}

\textcolor{black}{\textbf{Image:}}

\begin{center}
  \includegraphics[width=0.4\linewidth]{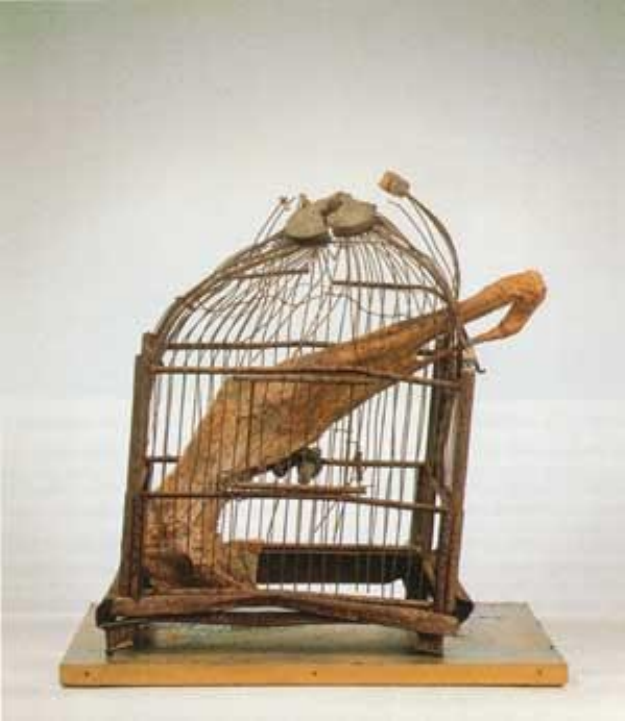}
\end{center}

\colorbox{gray!70}{%
    \parbox{\dimexpr\textwidth-2\fboxsep}{%
    \centering
        \textbf{Step 1}%
    }%
}
\textcolor{purple1}{\textbf{$<$think$>$} \\
I need to perform image search on the image in the question to obtain more information.\\
\textbf{$</$think$>$}}\\
\textcolor{black}{\textbf{$<$tool\_call$>$} \\ 
\{``name'': ``image\_search'', ``arguments'': \{\}\}
\\
\textbf{$</$tool\_call$>$}}\\
\textcolor{black}{
\textbf{$<$tool\_response$>$}\\
\begin{center}
  \includegraphics[width=0.2\linewidth]{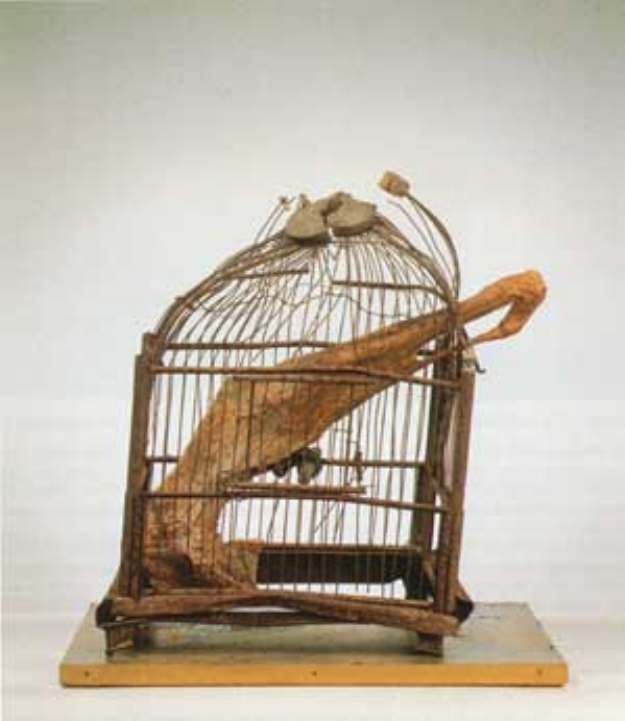}
\end{center}
1. [Page Link] https://www.wikiart.org/en/daniel-spoerri/cage-d-oiseau-1969 [Page Title] Cage d'oiseau, 1969 - Daniel Spoerri - WikiArt.org\\
\textbf{$</$tool\_response$>$}
}\\
\colorbox{gray!70}{%
    \parbox{\dimexpr\textwidth-2\fboxsep}{%
    \centering
        \textbf{Step 2}%
    }%
}\\
\textcolor{purple1}{\textbf{$<$think$>$} \\
I need to determine if Daniel Spoerri's \"Cage d'oiseau\" is considered installation art. \\
\textbf{$</$think$>$}
}\\
\textcolor{black}{\textbf{$<$tool\_call$>$} \\ 
\{``name'': ``text\_search'', ``arguments'': \{``query'': ``Daniel Spoerri Cage d'oiseau installation art''\}\}
\\
\textbf{$</$tool\_call$>$}}\\
\textcolor{black}{\textbf{$<$tool\_response$>$}\\
1. [Link] https://www.wikiart.org/en/daniel-spoerri/cage-d-oiseau-1969 [Title] Cage d'oiseau, 1969 - Daniel Spoerri [Snippet] 'Cage d'oiseau' was created in 1969 by Daniel Spoerri in Neo-Dada style. Find more prominent pieces of installation at Wikiart.org – best ...
\\
2. [Link] https://www.wikiart.org/en/daniel-spoerri/all-works [Title] Daniel Spoerri - 27 artworks - printmaking [Snippet] installation (20); performance (1); sculpture (4). Media. ready-made (1) ... Snare ... Cage d'oiseau - Daniel Spoerri. Cage d'oiseau Daniel Spoerri • 1969.
\\
3. [Link] https://esthete.org/bg/daniel-spoerri/the-boxes-snare-picture-1961 [Title] The Boxes (Snare-Picture), 1961 - Daniel Spoerri [Snippet] Explore Installation Art: The Boxes (Snare-Picture), 1961 by Daniel Spoerri ... Cage d'oiseau · Daniel Spoerri 1969. Veiw all 25 Artworks. Add to exposition
\\
4. [Link] https://www.artsy.net/artist/daniel-spoerri/auction-results [Title] Daniel Spoerri - Auction Results and Sales Data | Artsy - Page 5 [Snippet] Daniel Spoerri is a Swiss-born artist known for his unconventional and diverse artistic practice. Initially recognized as a dancer and performance artist, ...
\\
5. [Link] https://www.getty.edu/research/collections/component/10PHBK [Title] Series I. Events and artist projects, 1957-1990s - Getty Museum [Snippet] Daniel Spoerri artwork, various, 1960, 1963-1964. Viewable digital material ... Ann Weaver Norton, sculpture installation, home and sculpture garden, Florida, ...\\
\textbf{$</$tool\_response$>$}}\\
\colorbox{gray!70}{%
    \parbox{\dimexpr\textwidth-2\fboxsep}{%
    \centering
        \textbf{Step 3}%
    }%
}\\
\textcolor{purple1}{\textbf{$<$think$>$} \\
I need to determine if Marcel Duchamp's Bicycle Wheel is considered installation art. \\
\textbf{$</$think$>$}} \\
\textcolor{black}{\textbf{$<$tool\_call$>$} \\ 
\{``name'': ``text\_search'', ``arguments'': \{``query'': ``Marcel Duchamp Bicycle Wheel installation art''\}\}
\\
\textbf{$</$tool\_call$>$}}\\
\textcolor{black}{\textbf{$<$tool\_response$>$}\\
1. [Link] https://www.moma.org/collection/works/81631 [Title] Marcel Duchamp. Bicycle Wheel. New York, 1951 (third version ... [Snippet] Marcel Duchamp. Bicycle Wheel. New York, 1951 (third version, after lost original of 1913). Metal wheel mounted on painted wood stool.\\2. [Link] https://www.singulart.com/blog/en/2020/05/01/bicycle-wheel-1913-the-story-of-marcel-duchamps-pioneering-style/ [Title] Bicycle Wheel (1913): The Story of Marcel Duchamp's Pioneering ... [Snippet] May 1, 2020 ... The readymades were appropriated everyday objects, which Duchamp used to question the notion of art and to remove the notions of adoration and...\\3. [Link] https://www.moma.org/audio/playlist/296/3787 [Title] Marcel Duchamp. Bicycle Wheel. New York, 1951 (third version ... [Snippet] Ann Temkin: The whole idea of beauty as the goal, or simple enjoyment as the goal of art, he wanted to replace with this sense of art as something that provokes...\\4. [Link] https://en.wikipedia.org/wiki/Marcel\_Duchamp [Title] Marcel Duchamp - Wikipedia [Snippet] Exhibition design and installation art. 12 Étant donnés; 13 Personal life ... In 1913, Duchamp installed a Bicycle Wheel in his studio. The Bicycle Wheel...\\5. [Link] https://gagosian.com/exhibitions/2014/marcel-duchamp/ [Title] Marcel Duchamp, 980 Madison Avenue, New York, June 26–August ... [Snippet] Installation Views. Installation video. Installation video. Works Exhibited. Marcel Duchamp, Bicycle Wheel, 1913/64. (“Ex Arturo,”. Marcel Duchamp Bicycle Wheel...\\
\textbf{$</$tool\_response$>$}}\\
\colorbox{gray!70}{%
    \parbox{\dimexpr\textwidth-2\fboxsep}{%
    \centering
        \textbf{Step 4}%
    }%
}\\
\textcolor{purple1}{\textbf{$<$think$>$} \\
I need to compare Daniel Spoerri's \"Cage d'oiseau\" and Marcel Duchamp's \"Bicycle Wheel\" in terms of their classification as installation art.
\\
\textbf{$</$think$>$}} \\
\textcolor{black}{\textbf{$<$tool\_call$>$}\\
\{``name'': ``text\_search'', ``arguments'': \{``query'': ``Daniel Spoerri Cage d'oiseau vs Marcel Duchamp Bicycle Wheel installation art''\}\}\\
\textbf{$</$tool\_call$>$}}\\
\textcolor{black}{\textbf{$<$tool\_response$>$}\\
1. [Link] https://www.wikiart.org/en/daniel-spoerri/cage-d-oiseau-1969 [Title] Cage d'oiseau, 1969 - Daniel Spoerri [Snippet] 'Cage d'oiseau' was created in 1969 by Daniel Spoerri in Neo-Dada style. Find more prominent pieces of installation at Wikiart.org – best ...\\2. [Link] https://www.wikiart.org/en/daniel-spoerri/all-works [Title] Daniel Spoerri - 27 artworks - printmaking [Snippet] Eaten by Marcel Duchamp - Daniel Spoerri. Eaten by Marcel Duchamp Daniel ... Cage d'oiseau Daniel Spoerri • 1969. Brotteigobjekt Schreibmaschine - Daniel ...\\3. [Link] https://nikidesaintphalle.org/blog/ [Title] Blog [Snippet] Tinguely and Aeppli both became part of the avant-garde art scene in Paris where they first moved with another Swiss artist, Daniel Spoerri.\\4. [Link] https://monoskop.org/Art [Title] Art [Snippet] The page brings together some 350 art styles and movements from the 1860s to the present day, as they have come to us through these four paths.\\5. [Link] https://ilab.org/assets/catalogues [Title] ars libri ltd [Snippet] Lebel, with much assistance from Marcel Duchamp, and with Patrick Waldberg ... consisted of a bicycle wheel and some signs hanging from clothes-lines ......\\
\textbf{$</$tool\_response$>$}}\\
\colorbox{gray!70}{%
    \parbox{\dimexpr\textwidth-2\fboxsep}{%
    \centering
        \textbf{Step 5}%
    }%
}\\
\textcolor{purple1}{\textbf{$<$think$>$} \\
I need to determine the classification of Daniel Spoerri's Cage d'oiseau as either installation art, sculpture, or ready-made.\\
\textbf{$</$think$>$}} \\
\textcolor{black}{\textbf{$<$tool\_call$>$} \\ 
\{``name'': ``text\_search'', ``arguments'': \{``query'': ``Daniel Spoerri Cage d'oiseau classification installation art sculpture ready-made''\}\}\\
\textbf{$</$tool\_call$>$}}\\
\textcolor{black}{\textbf{$<$tool\_response$>$}\\
1. [Link] https://www.wikiart.org/en/daniel-spoerri/cage-d-oiseau-1969 [Title] Cage d'oiseau, 1969 - Daniel Spoerri [Snippet] 'Cage d'oiseau' was created in 1969 by Daniel Spoerri in Neo-Dada style. Find more prominent pieces of installation at Wikiart.org – best ...\\2. [Link] https://www.wikiart.org/en/daniel-spoerri/all-works [Title] Daniel Spoerri - 27 artworks - printmaking [Snippet] ready-made (1) ... Snare Picture - Daniel Spoerri. Snare Picture Daniel Spoerri ... Cage d'oiseau - Daniel Spoerri. Cage d'oiseau Daniel Spoerri • 1969.\\3. [Link] https://nikidesaintphalle.org/blog/ [Title] Blog [Snippet] Tinguely and Aeppli both became part of the avant-garde art scene in Paris where they first moved with another Swiss artist, Daniel Spoerri.\\4. [Link] https://www.artsy.net/artist/daniel-spoerri [Title] Daniel Spoerri - Biography, Shows, Articles More [Snippet] Daniel Spoerri is a Swiss-born artist known for his unconventional and diverse artistic practice. Initially recognized as a dancer and performance artist.\\5. [Link] https://www.artedio.com/daniel-spoerri/artworks-editions [Title] Daniel Spoerri Artworks And Editions For Sale - ARTEDIO [Snippet] Daniel Spoerri artworks and limited editions for sale at online art gallery ARTEDIO. Buy Daniel Spoerri prints and originals easily and safely online now....\\
\textbf{$</$tool\_response$>$}}\\
\colorbox{gray!70}{%
    \parbox{\dimexpr\textwidth-2\fboxsep}{%
    \centering
        \textbf{Step 6}%
    }%
}\\
\textcolor{purple1}{\textbf{$<$think$>$} \\
I need to determine if Marcel Duchamp's Bicycle Wheel is considered installation art.\\
\textbf{$</$think$>$}} \\
\textcolor{black}{\textbf{$<$tool\_call$>$} \\ 
\{``name'': ``text\_search'', ``arguments'': \{``query'': ``Is Marcel Duchamp's Bicycle Wheel installation art?''\}\}\\
\textbf{$</$tool\_call$>$}}\\
\textcolor{black}{\textbf{$<$tool\_response$>$}\\
1. [Link] https://www.moma.org/collection/works/81631 [Title] Marcel Duchamp. Bicycle Wheel. New York, 1951 (third version ... [Snippet] Though Bicycle Wheel predates Marcel Duchamp's use of the word “readymade”—a term he coined after moving from Paris to New York in 1915—it is the earliest ...\\2. [Link] https://www.singulart.com/blog/en/2020/05/01/bicycle-wheel-1913-the-story-of-marcel-duchamps-pioneering-style/ [Title] Bicycle Wheel (1913): The Story of Marcel Duchamp's Pioneering ... [Snippet] May 1, 2020 ... The readymades were appropriated everyday objects, which Duchamp used to question the notion of art and to remove the notions of adoration and ...\\3. [Link] https://www.moma.org/audio/playlist/296/3787 [Title] Marcel Duchamp. Bicycle Wheel. New York, 1951 (third version ... [Snippet] Artist, Marcel Duchamp (archival): The first one was in 1913. It was a bicycle wheel. Curator, Ann Temkin: This is Duchamp himself talking about Bicycle ...\\4. [Link] https://en.wikipedia.org/wiki/Bicycle\_Wheel [Title] Bicycle Wheel - Wikipedia [Snippet] Bicycle Wheel is a readymade from Marcel Duchamp consisting of a bicycle fork with front wheel mounted upside-down on a wooden stool.\\5. [Link] https://gagosian.com/exhibitions/2014/marcel-duchamp/ [Title] Marcel Duchamp, 980 Madison Avenue, New York, June 26–August ... [Snippet] Installation Views. Installation video. Installation video. Works Exhibited. Marcel Duchamp, Bicycle Wheel, 1913/64. (“Ex Arturo,”. Marcel Duchamp Bicycle Wheel ...\\
\textbf{$</$tool\_response$>$}}\\

\colorbox{gray!70}{%
    \parbox{\dimexpr\textwidth-2\fboxsep}{%
    \centering
        \textbf{Step 7}%
    }%
}\\
\textcolor{purple1}{\textbf{$<$think$>$} \\
I need to determine if Daniel Spoerri's Cage d'oiseau is considered installation art.\\
\textbf{$</$think$>$}} \\
\textcolor{black}{\textbf{$<$tool\_call$>$} \\ 
\{``name'': ``text\_search'', ``arguments'': \{``query'': ``Is Daniel Spoerri's Cage d'oiseau installation art?''\}\}\\
\textbf{$</$tool\_call$>$}}\\
\textcolor{black}{\textbf{$<$tool\_response$>$}\\
1. [Link] https://www.wikiart.org/en/daniel-spoerri/cage-d-oiseau-1969 [Title] Cage d'oiseau, 1969 - Daniel Spoerri [Snippet] 'Cage d'oiseau' was created in 1969 by Daniel Spoerri in Neo-Dada style. Find more prominent pieces of installation at Wikiart.org – best ...\\2. [Link] https://www.wikiart.org/en/daniel-spoerri/all-works [Title] Daniel Spoerri - 27 artworks - printmaking [Snippet] installation (20); performance (1); sculpture (4). Media. ready-made (1) ... Snare ... Cage d'oiseau - Daniel Spoerri. Cage d'oiseau Daniel Spoerri • 1969.\\3. [Link] https://esthete.org/bg/daniel-spoerri/the-boxes-snare-picture-1961 [Title] The Boxes (Snare-Picture), 1961 - Daniel Spoerri [Snippet] Explore Installation Art: The Boxes (Snare-Picture), 1961 by Daniel Spoerri ... Cage d'oiseau · Daniel Spoerri 1969. Veiw all 25 Artworks. Add to exposition\\4. [Link] https://www.artsy.net/artist/daniel-spoerri/auction-results [Title] Daniel Spoerri - Auction Results and Sales Data | Artsy - Page 5 [Snippet] Daniel Spoerri is a Swiss-born artist known for his unconventional and diverse artistic practice. Initially recognized as a dancer and performance artist, ...\\5. [Link] https://nikidesaintphalle.org/blog/ [Title] Blog [Snippet] Tinguely and Aeppli both became part of the avant-garde art scene in Paris where they first moved with another Swiss artist, Daniel Spoerri.\\
\textbf{$</$tool\_response$>$}}\\

\colorbox{gray!70}{%
    \parbox{\dimexpr\textwidth-2\fboxsep}{%
    \centering
        \textbf{Step 8}%
    }%
}\\
\textcolor{purple1}{\textbf{$<$think$>$} \\
I have gathered enough information. Now I am confident to answer the question.\\
\textbf{$</$think$>$}} \\

\textcolor{red}{\textbf{$<$answer$>$}Yes.
\textbf{$</$answer$>$}} \textcolor{black}{\checkmark}\\

\end{tcolorbox}


\end{document}